# Enhancing QPNs for Trade-off Resolution

**Silja Renooij** and **Linda C. van der Gaag**
Utrecht University, Department of Computer Science
P.O. Box 80.089, 3508 TB Utrecht, The Netherlands
{silja,linda}@cs.uu.nl

## Abstract

Qualitative probabilistic networks have been introduced as qualitative abstractions of Bayesian belief networks. One of the major drawbacks of these qualitative networks is their coarse level of detail, which may lead to unresolved trade-offs during inference. We present an enhanced formalism for qualitative networks with a finer level of detail. An enhanced qualitative probabilistic network differs from a regular qualitative network in that it distinguishes between strong and weak influences. Enhanced qualitative probabilistic networks are purely qualitative in nature, as regular qualitative networks are, yet allow for efficiently resolving trade-offs during inference.

## 1 INTRODUCTION

The formalism of Bayesian belief networks is generally considered an intuitively appealing and powerful formalism for capturing complex problem domains along with their uncertainties. The usually large number of probabilities required for a belief network, however, tends to pose a major obstacle to their application. To mitigate this obstacle, *qualitative probabilistic networks* have been introduced as qualitative abstractions of Bayesian belief networks [Wellman, 1990]. Like a Bayesian belief network, a qualitative probabilistic network encodes variables and the probabilistic interrelationships among these variables in a directed graph; the relationships are not quantified by conditional probabilities as in a belief network, but are summarised by qualitative signs instead. For inference with a qualitative probabilistic network, an elegant algorithm is available, based on the idea of propagating signs [Druzdzel & Henrion, 1993].

One of the major drawbacks of qualitative probabilistic networks is their coarse level of detail. As a consequence of their high abstraction level, qualitative probabilistic networks do not provide for modelling the intricacies involved in weighing conflicting influences and, hence, do not provide for resolving trade-offs. Inference with a qualitative probabilistic network for a real-life domain of application, therefore, quite often leads to ambiguous results.

Ambiguous results in inference can be averted by enhancing the formalism of qualitative probabilistic networks to provide for a finer level of detail. Roughly speaking, the finer the level of detail, the more trade-offs can be resolved during inference. The problem of trade-off resolution within the framework of qualitative networks has been addressed before by others. S. Parsons has introduced, for example, the concept of categorical influences. A categorical influence is either an influence that serves to increase a probability to 1 or an influence that decreases a probability to 0, regardless of any other influences, and thereby resolves any trade-off in which it is involved [Parsons, 1995]. C.-L. Liu and M.P. Wellman have designed two methods for resolving trade-offs based upon the idea of reverting to numerical probabilities whenever necessary [Liu & Wellman, 1998]. While only some trade-offs can be resolved by the use of categorical influences, the methods of Liu and Wellman provide for resolving any trade-off. Their methods, however, require a fully specified, numerical belief network. We would like to mention that various other approaches to dealing with uncertainty in a qualitative way have been proposed in the literature. These approaches are not tailored for use within the framework of qualitative probabilistic networks and therefore will not be reviewed here.

To provide for trade-off resolution without resorting to numerical probabilistic information, we have designed an intuitively appealing formalism of *enhanced qualitative networks*. An enhanced qualitative probabilistic network differs from a regular qualitative network in that it distinguishes between strong and weak influences. For inference, we have generalised the sign-propagation algorithm for regular qualitative networks to deal with the strong and weak influences of an enhanced qualitative network. Trade-off resolution during inference is based on the idea that strong influences dominate over conflicting weak influences.

The paper is organised as follows. In Section 2, we provide some preliminaries from the field of qualitative networks to introduce our notational conventions. In Section 3, we present the formalism of enhanced qualitative probabilistic networks. In Section 4, we detail various properties of these enhanced networks, thereby providing for a sign-propagation algorithm for inference. The paper is rounded off with some conclusions and directions for future research in Section 6.

## 2  PRELIMINARIES

*Qualitative probabilistic networks* have been introduced as abstractions of Bayesian belief networks. Before addressing qualitative networks, we briefly review their quantitative counterparts. A Bayesian belief network is a concise representation of a joint probability distribution on a set of statistical variables. It encodes, in an acyclic directed graph, the variables concerned along with their probabilistic interrelationships. Each node in the digraph represents a variable; the probabilistic relationships between the variables are captured in the digraph's set of arcs. Associated with each variable is a set of conditional probability distributions describing the relationship of this variable with its (immediate) predecessors in the digraph.

We introduce a small Bayesian belief network that will serve as our running example throughout the paper.

**Example 2.1** We consider the small belief network shown in Figure 1. The network represents a fragment of fictitious and incomplete medical knowledge, pertaining to the effects of administering antibiotics on a patient. Node $A$ represents whether or not a patient takes antibiotics. Node $T$ models whether or not the patient has typhoid fever and node $D$ represents presence or absence of diarrhoea in the patient. Node $F$, to conclude, describes whether or not the composition of the patient's bacterial flora has changed. Typhoid fever and a change in the patient's bacterial flora are modelled as the possible causes of diarrhoea. Antibiotics can cure typhoid fever by killing the bacteria that cause the infection. However, antibiotics can also change the composition of the patient's bacterial flora, thereby increasing the risk of diarrhoea. □

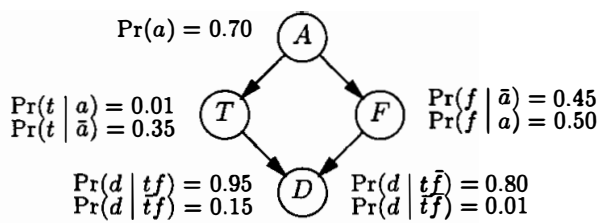

Figure 1: The *Antibiotics* Belief Network.

Qualitative probabilistic networks bear a strong resemblance to their quantitative counterparts. A qualitative probabilistic network also comprises an acyclic digraph modelling variables and probabilistic interrelationships among variables. Instead of conditional probability distributions, however, a qualitative probabilistic network associates with its digraph *qualitative influences* and *qualitative synergies* [Wellman, 1990].

A qualitative influence between two nodes expresses how the values of one node influence the probabilities of the values of the other node. A *positive qualitative influence* of node $A$ on its (immediate) successor $B$, denoted $S^+(A,B)$, expresses that observing higher values for $A$ makes higher values for $B$ more likely, regardless of any other direct influence on $B$, that is,

$$\Pr(b \mid ax) - \Pr(b \mid \bar{a}x) \geq 0$$

for any combination of values $x$ for the set $\pi(B) \setminus \{A\}$ of (immediate) predecessors of $B$ other than $A$. A *negative qualitative influence*, denoted by $S^-$, and a *zero qualitative influence*, denoted by $S^0$, are defined analogously, replacing $\geq$ in the above formula by $\leq$ and $=$, respectively. If the influence of node $A$ on node $B$ is not monotonic or unknown, we say that it is *ambiguous*, denoted $S^?(A,B)$.

The set of influences of a qualitative probabilistic network exhibits various convenient properties [Wellman, 1990]. The property of *symmetry* guarantees that, if the network includes the influence $S^+(A,B)$, then it also includes $S^+(B,A)$. The property of *transitivity* asserts that qualitative influences along a trail, that specifies at most one incoming arc for each node, combine into a single influence with the ⊗-operator from Figure 2. The property of *composition* asserts that multiple qualitative influences between two nodes along parallel chains combine into a single influence with the ⊕-operator.

| ⊗ | + | − | 0 | ? |   | ⊕ | + | − | 0 | ? |
|---|---|---|---|---|---|---|---|---|---|---|
| + | + | − | 0 | ? |   | + | + | ? | + | ? |
| − | − | + | 0 | ? |   | − | ? | − | − | ? |
| 0 | 0 | 0 | 0 | 0 |   | 0 | + | − | 0 | ? |
| ? | ? | ? | 0 | ? |   | ? | ? | ? | ? | ? |

Figure 2: The ⊗- and ⊕-Operators.

From Figure 2, we have that combining parallel qualitative influences with the ⊕-operator may yield an ambiguous result. Such an ambiguity, in fact, results whenever influences with opposite signs are combined. We say that the *trade-off* that is reflected by the conflicting influences cannot be *resolved*. Note that, in contrast with the ⊕-operator, the ⊗-operator cannot introduce ambiguities upon combining signs of influences along trails.

In addition to influences, a qualitative probabilistic network includes synergies, that express how the value of one node influences the probabilities of the values of another node in view of a given value for a third node [Henrion & Druzdzel, 1991]. A *negative product synergy* of node $A$ on node $B$ (and vice versa) given the value $c$ for their common successor $C$, denoted $X^-(\{A,B\},c)$, expresses that, given $c$, higher values for $A$ render higher values for $B$ less likely, that is,

$$\Pr(c \mid abx)\cdot\Pr(c \mid \bar{a}\bar{b}x) - \Pr(c \mid a\bar{b}x)\cdot\Pr(c \mid \bar{a}bx) \leq 0$$

for any combination of values $x$ for the set $\pi(C) \setminus \{A,B\}$ of predecessors of $C$ other than $A$ and $B$. A product synergy induces a qualitative influence between the predecessors of a node upon observation; the induced influence is coined an *intercausal influence*. *Positive*, *zero*, and *ambiguous product synergies* again are defined analogously.

**Example 2.2** We consider the qualitative abstraction of the *Antibiotics* belief network from Figure 1. From the conditional probability distributions specified for node $T$, we have that

$$\Pr(t \mid a) - \Pr(t \mid \bar{a}) \leq 0$$

and therefore that $S^-(A,T)$; we further find that $S^+(A,F)$, $S^+(T,D)$, and $S^+(F,D)$. Either value for node $D$, in addition, induces a negative intercausal influence between the nodes $T$ and $F$. The resulting qualitative probabilistic network is shown in Figure 3. □

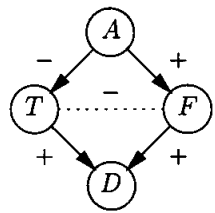

Figure 3: The Qualitative *Antibiotics* Network.

We would like to note that, although in the previous example, we have computed the qualitative probabilistic relationships among the variables from the probabilities of the original belief network, in real-life applications, these relationships are elicited directly from domain experts.

For reasoning with a qualitative probabilistic network, an elegant algorithm is available from M.J. Druzdzel and M. Henrion (1993); this algorithm is summarised in pseudocode in Figure 4. The basic idea of the algo-

**procedure** Propagate-Sign(*from,to,message*):
sign[*to*] ← sign[*to*] ⊕ *message*;
**for** each (induced) neighbour $V_i$ of *to*
**do** *linksign* ← sign of (induced) influence
            between *to* and $V_i$;
   *message* ← sign[*to*] ⊗ *linksign*;
   **if** $V_i \neq from$ and $V_i \notin Observed$
      and sign[$V_i$] ≠ sign[$V_i$] ⊕ *message*
   **then** Propagate-Sign(*to,$V_i$,message*)

Figure 4: The Sign-Propagation Algorithm.

rithm is to trace the effect of observing a node's value on the other nodes in the network by message-passing between neighbouring nodes. For each node, a *sign* is determined, indicating the direction of change in the node's probabilities occasioned by the new observation given all previously observed node values. Initially, all node signs equal '0'. For the newly observed node, an appropriate sign is entered, that is, either a '+' for the observed value *true* or a '−' for the value *false*. The node updates its sign and subsequently sends a message to each neighbour and every node on which it exerts an induced intercausal influence. The sign of this message is the ⊗-product of the node's (new) sign and the sign of the influence it traverses. This process is repeated throughout the network, building on the properties of symmetry, transitivity, and composition of influences.

## 3 THE ENHANCED FORMALISM

Qualitative probabilistic networks model a problem domain at a coarse level of detail. This coarseness of representation is most visible in the way relationships among variables are captured: the relationships are summarised by qualitative influences without any indication of their strengths. As a consequence of the coarse level of detail, any trade-off encountered during inference will remain unresolved. To allow for resolving trade-offs in a qualitative way, we enhance the formalism of qualitative probabilistic networks by associating a relative strength with influences. If in a trade-off, for example, the positive influence is known to be stronger than the conflicting negative one, we may then conclude the combined influence to be positive, thereby resolving the trade-off.

In our formalism of *enhanced qualitative probabilistic networks*, we distinguish between strong and weak influences. We begin by focusing on the strong and weak positive influences. The basic idea is to partition the set of all positive influences into two disjoint sets of influences in such a way that any influence from the one subset is stronger than any influence from the other subset. To this end, a *cut-off value* $\delta$ is introduced. This value serves to partition the set of qualitative influences into a set of influences that capture a difference in probabilities larger than $\delta$ and a set of influences that model a difference smaller than $\delta$. An influence from the former subset will be termed a strongly positive influence; an influence from the latter subset will be termed a weakly positive influence.

More formally, a *strongly positive qualitative influence* of a node $A$ on its successor $B$, denoted $S^{++}(A,B)$, expresses, first and foremost, that observing higher values for $A$ makes higher values for $B$ more likely, regardless of any other influence on $B$; in addition, it expresses that

$$\Pr(b \mid ax) - \Pr(b \mid \bar{a}x) \geq \delta$$

for any combination of values $x$ for the set $\pi(B) \setminus \{A\}$ of predecessors of $B$ other than $A$, where $\delta$ is the cut-off value used. A *weakly positive qualitative influence* of $A$ on $B$, denoted $S^{+}(A,B)$, is a positive qualitative influence such that

$$\Pr(b \mid ax) - \Pr(b \mid \bar{a}x) \leq \delta$$

for any combination of values $x$ for the set $\pi(B) \setminus \{A\}$ of predecessors of $B$ other than $A$, where $\delta$ once again is the cut-off value used. *Strongly negative qualitative influences*, denoted $S^{--}$, and *weakly negative qualitative influences*, denoted $S^{-}$, are defined analogously; *zero qualitative influences* and *ambiguous qualitative influences* are defined as in regular qualitative probabilistic networks. In the sequel, we will use the phrase *strong influences* to refer to both strongly positive and strongly negative influences; the phrase *weak influences* is meant to have an analogous meaning. We further say that a product synergy is *strongly negative* if it induces a strongly negative intercausal influence. *Strongly positive product synergies* are defined analogously; *zero product synergies* and *ambiguous product synergies* again are defined as in regular qualitative networks.

We would like to note that, in our enhanced formalism, the meaning of the sign of an influence has slightly changed. While in a regular qualitative probabilistic network, the sign of an influence represents the sign of a difference in probabilities only, in an enhanced qualitative network a sign in addition captures the relative magnitude of the difference.

Upon abstracting a Bayesian belief network to an enhanced qualitative probabilistic network, the cut-off value $\delta$ needs to be chosen explicitly. This cut-off value will typically vary from application to application. Note that it is always possible to choose a cut-off value, as the value $\delta = 1$ yields a trivial partitioning of the set of influences.

**Example 3.1** We consider once again the *Antibiotics* belief network from Example 2.1. Suppose that we choose for our cut-off value $\delta = 0.30$. For the influence of node $A$ on node $T$, we now find that

$$\Pr(t \mid a) - \Pr(t \mid \bar{a}) \leq 0, \text{ and}$$
$$|\Pr(t \mid a) - \Pr(t \mid \bar{a})| = 0.34 \geq \delta$$

We therefore conclude that $S^{--}(A,T)$. We further find that $S^{++}(T,D)$, $S^{+}(A,F)$, and $S^{+}(F,D)$. The resulting enhanced qualitative probabilistic network is shown in Figure 5. $\square$

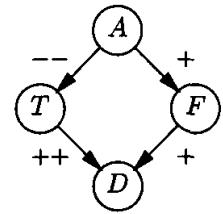

Figure 5: The Enhanced *Antibiotics* Network.

We would like to note that, in real-life applications of enhanced qualitative probabilistic networks, a cut-off value need not be established explicitly. The partitioning into strong and weak influences then is elicited directly from the domain experts involved in the construction of the network.

## 4   INFERENCE WITH AN ENHANCED NETWORK

For inference with a regular qualitative probabilistic network, an elegant algorithm is available. We recall from Section 2 that this algorithm builds on the idea of propagating signs throughout a network and combining them with the $\otimes$- and $\oplus$-operators. We further recall that the algorithm exploits the properties of symmetry, transitivity, and parallel composition of influences. To generalise the idea of sign-propagation to inference with an enhanced qualitative probabilistic network, we enhance, in the Sections 4.1 and 4.2, the $\otimes$- and $\oplus$-operators to provide for the properties of transitivity and parallel combination of strong and weak influences; in Section 4.3, we address the property of symmetry.

### 4.1 ENHANCING THE ⊗-OPERATOR

For propagating qualitative signs along trails of nodes in an enhanced qualitative probabilistic network, we enhance the ⊗-operator that is defined for regular qualitative networks, to apply to strong and weak influences. We recall that the ⊗-operator basically provides for multiplying signs of influences. In a regular qualitative probabilistic network, an influence captures a difference between two probabilities. Upon multiplying the signs of two influences, therefore, the sign of the result of the multiplication of two such differences is computed. In our formalism of enhanced qualitative probabilistic networks, we have added an explicit notion of relative magnitude to influences. It will be evident that these relative magnitudes need to be taken into consideration when multiplying signs.

To address the effect of multiplying two signs in an enhanced qualitative probabilistic network, we consider the network fragment shown in Figure 6. The frag-ment includes the trail of nodes $A$, $B$, $C$, with two qualitative influences between them; in addition, $X$ denotes the set of all predecessors of $B$ other than $A$, and $Y$ is the set of all predecessors of $C$ other than $B$. For the qualitative influence of $A$ on $C$, we have that

$$\Pr(c \mid axy) - \Pr(c \mid \bar{a}xy) =$$
$$(\Pr(c \mid by) - \Pr(c \mid \bar{b}y)) \cdot (\Pr(b \mid ax) - \Pr(b \mid \bar{a}x))$$

for any combination of values $x$ for the set of nodes $X$ and any combination of values $y$ for the set $Y$.

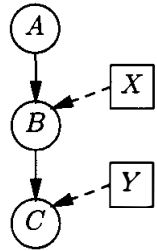

Figure 6: A Fragment of a Network.

Suppose that both qualitative influences in the network fragment under consideration are strongly positive, that is, we have that $S^{++}(A, B)$ and $S^{++}(B, C)$; suppose that we have used the cut-off value $\delta$ for distinguishing between strong and weak influences. From the expression stated above for the influence of node $A$ on node $C$, we now find that

$$\Pr(c \mid axy) - \Pr(c \mid \bar{a}xy) \geq \delta^2$$

for any combination of values $xy$ for the set of nodes $X \cup Y$. Since $\delta \leq 1$, we have that $\delta^2 \leq \delta$. Upon multiplying the signs of two strong influences, therefore, a sign results that expresses an influence that may or may not be stronger than a single weakly positive influence.

Now suppose that both qualitative influences in the network fragment from Figure 6 are weakly positive, that is, we have that $S^+(A, B)$ and $S^+(B, C)$. For the influence of node $A$ on node $C$, we now find that

$$\Pr(c \mid axy) - \Pr(c \mid \bar{a}xy) \leq \delta^2$$

for any combination of values $xy$ for the set $X \cup Y$. While the influence resulting from the multiplication of two strong influences cannot be compared to a single weak influence, the above observation shows that the resulting influence will always be at least as strong as an influence resulting from the multiplication of two weak influences. To provide for comparing qualitative influences along different trails with respect to their magnitude, as required for trade-off resolution, therefore, we need to retain the length of the trail in the network over which influences have been multiplied.

To provide for comparing qualitative influences along different trails, we augment every influence's sign by a superscript, called the sign's *multiplication index*. A strongly positive qualitative influence with multiplication index $i$ of node $A$ on node $B$, written $S^{++^i}(A, B)$, is now taken to denote that

$$\Pr(b \mid ax) - \Pr(b \mid \bar{a}x) \geq \delta^i$$

for every combination of values $x$ for the set $X$ of predecessors of $B$ other than $A$. A weakly positive qualitative influence with multiplication index $i$ of $A$ on $B$, written $S^{+^i}(A, B)$, is taken to indicate that

$$0 \leq \Pr(b \mid ax) - \Pr(b \mid \bar{a}x) \leq \delta^i$$

for every combination of values $x$ for the set $X$. The signs associated with the arcs of the digraph are interpreted as having a multiplication index equal to 1.

Building on the concept of multiplication index, Figure 7 shows the table for the enhanced ⊗-operator. From the table, it is readily seen that the +, −, 0, and ? signs combine as in a regular qualitative probabilistic network; the difference is just in the handling of the multiplication indices. In the table, there appear signs $+_?$ and $-_?$; we will elaborate on the meaning of these signs in Section 4.2.

We like to further comment on the combination of the signs $+^i$ and $++^j$. In doing so, we consider once again the network fragment from Figure 6. Suppose that we have $S^{+^i}(A, B)$ for the influence of node $A$ on node $B$, and $S^{++^j}(B, C)$ for the influence of $B$ on $C$. The weakly positive influence of $A$ on $B$ expresses that

$$\Pr(b \mid ax) - \Pr(b \mid \bar{a}x) \leq \delta^i$$

| $\otimes$ | $++^j$ | $+^j$ | $+?$ | $0$ | $-?$ | $-^j$ | $--^j$ | $?$ |
|---|---|---|---|---|---|---|---|---|
| $++^i$ | $++^{i+j}$ | $+^j$ | $+?$ | $0$ | $-?$ | $-^j$ | $--^{i+j}$ | $?$ |
| $+^i$ | $+^i$ | $+^{i+j}$ | $+^i$ | $0$ | $-^i$ | $-^{i+j}$ | $-^i$ | $?$ |
| $+?$ | $+?$ | $+^j$ | $+?$ | $0$ | $-?$ | $-^j$ | $-?$ | $?$ |
| $0$ | $0$ | $0$ | $0$ | $0$ | $0$ | $0$ | $0$ | $0$ |
| $-?$ | $-?$ | $-^j$ | $-?$ | $0$ | $+?$ | $+^j$ | $+?$ | $?$ |
| $-^i$ | $-^i$ | $-^{i+j}$ | $-^i$ | $0$ | $+^i$ | $+^{i+j}$ | $+^i$ | $?$ |
| $--^i$ | $--^{i+j}$ | $-^j$ | $-?$ | $0$ | $+?$ | $+^j$ | $++^{i+j}$ | $?$ |
| $?$ | $?$ | $?$ | $?$ | $0$ | $?$ | $?$ | $?$ | $?$ |

Figure 7: The Enhanced $\otimes$-Operator.

for every combination of values $x$ for the set $X$ of predecessors of $B$ other than $A$. The strongly positive qualitative influence of $B$ on $C$ further expresses that

$$\Pr(c \mid by) - \Pr(c \mid \bar{b}y) \geq \delta^j$$

for every combination of values $y$ for the set $Y$ of predecessors of $C$ other than $B$. For the influence of $A$ on $C$, we now find that

$$\Pr(c \mid axy) - \Pr(c \mid \bar{a}xy) \leq \delta^i$$

for every combination of values $xy$ for the set $X \cup Y$. We therefore conclude that $S^{+^i}(A, C)$. So, $+^i \otimes ++^j = +^i$. Similar observations apply to any multiplication of a weak and a strong influence.

### 4.2 ENHANCING THE $\oplus$-OPERATOR

For combining multiple qualitative influences between two nodes along parallel trails in an enhanced qualitative network, we enhance the $\oplus$-operator that is defined for regular qualitative probabilistic networks, to apply to strong and weak influences. We recall that the $\oplus$-operator basically provides for adding signs of influences. We further recall that, upon adding the signs of two conflicting influences in a regular qualitative network, the represented trade-off cannot be resolved and an ambiguous influence results. In our formalism of enhanced qualitative probabilistic networks, we have added an explicit notion of relative magnitude to influences. These relative magnitudes can now be taken into consideration when adding the signs of conflicting influences and used to resolve trade-offs, thereby forestalling ambiguous results.

When addressing the enhanced $\otimes$-operator, in the previous section, we have argued that the multiplication of two influences yields an influence of possibly smaller magnitude. We will now see that the addition of two influences, in contrast, may result in an influence of larger magnitude. To address the effect of adding two signs in an enhanced qualitative probabilistic network, we consider the network fragment shown in Figure 8. The fragment includes the parallel trails $A$, $C$, and $A$, $B$, $C$, respectively, between the nodes $A$ and $C$, and various qualitative influences; in addition, $X$ denotes the set of all predecessors of $B$ other than $A$, and $Y$ is the set of all predecessors of $C$ other than $A$ and $B$. For the net qualitative influence of node $A$ on node $C$ along the two parallel trails, we have that

$$\Pr(c \mid axy) - \Pr(c \mid \bar{a}xy) =$$

$(\Pr(c \mid aby) - \Pr(c \mid a\bar{b}y)) \cdot \Pr(b \mid ax) +$
$-(\Pr(c \mid \bar{a}by) - \Pr(c \mid \bar{a}\bar{b}y)) \cdot \Pr(b \mid \bar{a}x) +$
$+(\Pr(c \mid a\bar{b}y) - \Pr(c \mid \bar{a}\bar{b}y))$

for any combination of values $x$ for the set of nodes $X$ and any combination of values $y$ for the set $Y$.

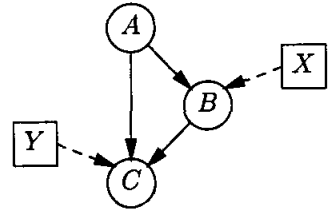

Figure 8: Another Network Fragment.

Suppose that all qualitative influences in the network fragment under consideration are weakly positive, that is, we have that $S^+(A, B)$, $S^+(B, C)$, and $S^+(A, C)$; suppose that we have used the cut-off value $\delta$ for distinguishing between strong and weak influences. The net influence of node $A$ on node $C$ equals the sum of the influence with sign $+^1$ along the trail $A$, $C$, and the influence with sign $+^2$ along the trail $A$, $B$, $C$. From the expression stated above for the net influence of $A$ on $C$, we find that

$$\Pr(c \mid axy) - \Pr(c \mid \bar{a}xy) \geq 0$$

The minimum of this difference is attained, for example, for $\Pr(c \mid a\bar{b}y) = 0$, which enforces $\Pr(c \mid \bar{a}\bar{b}y) = 0$, and $\Pr(b \mid ax) = \Pr(b \mid \bar{a}x) = 0$. We further find that

$$\Pr(c \mid axy) - \Pr(c \mid \bar{a}xy) \leq \delta + \delta^2$$

| $\oplus$ | $++^j$ | $+^j$ | $+_?$ | 0 | $-_?$ | $-^j$ | $--^j$ | ? |
|---|---|---|---|---|---|---|---|---|
| $++^i$ | $++^m$ | $++^i$ | $++^i$ | $++^i$ | ? | a) | ? | ? |
| $+^i$ | $++^j$ | $+_?$ | $+_?$ | $+^i$ | ? | ? | d) | ? |
| $+_?$ | $++^j$ | $+_?$ | $+_?$ | $+_?$ | ? | ? | ? | ? |
| 0 | $++^j$ | $+^j$ | $+_?$ | 0 | $-_?$ | $-^j$ | $--^j$ | ? |
| $-_?$ | ? | ? | ? | $-_?$ | $-_?$ | $-_?$ | $--^j$ | ? |
| $-^i$ | b) | ? | ? | $-^i$ | $-_?$ | $-_?$ | $--^j$ | ? |
| $--^i$ | ? | c) | ? | $--^i$ | $--^i$ | $--^i$ | $--^m$ | ? |
| ? | ? | ? | ? | ? | ? | ? | ? | ? |

where $m = min(i,j)$,

a) $+_?$, if $i \leq j$; ?, otherwise

b) $+_?$, if $j \leq i$; ?, otherwise

c) $-_?$, if $i \leq j$; ?, otherwise

d) $-_?$, if $j \leq i$; ?, otherwise

Figure 9: The Enhanced $\oplus$-Operator.

This maximum is attained, for example, for $\Pr(c \mid aby) = 1$, $\Pr(c \mid a\bar{b}y) = 1 - \delta$, $\Pr(c \mid \bar{a}by) = 1 - 2 \cdot \delta$, $\Pr(c \mid \bar{a}\bar{b}y) = 1 - \delta$, and $\Pr(b \mid ax) = 1$. In computing the maximum of the difference, we have used explicitly the information that all influences are weakly positive. From the maximum attained, it is readily seen that the addition of two weakly positive influences yields a result that may or may not be stronger than a weakly positive influence. In general, we have that the result of adding two positive or two negative influences is at least as strong as the strongest of the influences added.

From the preceding observations, we have that the qualitative influence that results from adding two weakly positive influences, is either weakly positive or strongly positive. So, although the resulting influence is known to be positive, its relative magnitude is unknown. To capture this ambiguity, we use $+_?$ to denote the influence's sign. An ambiguously positive qualitative influence of node $A$ on node $C$, written $S^{+?}(A, C)$, is therefore taken to indicate that

$$0 \leq \Pr(c \mid axy) - \Pr(c \mid \bar{a}xy) \leq 1$$

for any combination of values $xy$ for the set $X \cup Y$. Similarly, $-_?$ is used to denote an ambiguously negative qualitative influence.

The enhanced $\oplus$-operator is shown in Figure 9. From the table, it is readily seen that the $+$, $-$, 0, and ? signs combine as in a regular qualitative probabilistic network; the difference is just in the handling of the multiplication indices and the ambiguity subscripts.

We like to further comment on the resolution of trade-offs using the enhanced $\oplus$-operator. In doing so, we consider once again the network fragment from Figure 8. Suppose that we have $S^{++}(A, C)$ for the direct influence of node $A$ on node $C$, and that we further have $S^+(A, B)$ and $S^-(B, C)$. The net influence of node $A$ on node $C$ equals the sum of the influence with sign $++^1$ along the trail $A, C$, and the influence with sign $-^2$ along the trail $A, B, C$. From the expression for the net influence of $A$ on $C$, we find that

$$\Pr(c \mid axy) - \Pr(c \mid \bar{a}xy) \geq \delta - \delta^2$$

The minimum for the difference is attained, for example, for $\Pr(c \mid aby) = 2 \cdot \delta$, $\Pr(c \mid a\bar{b}y) = \delta$, $\Pr(c \mid \bar{a}by) = \delta$, $\Pr(c \mid \bar{a}\bar{b}y) = 0$, and $\Pr(b \mid ax) - \Pr(b \mid \bar{a}x) = \delta$. In computing the minimum of the difference, we have once again exploited the information with regard to the signs and relative magnitudes of the influences involved. From the minimum attained, it is readily seen that the net influence of node $A$ on node $C$ is positive. However, as $\delta - \delta^2 < \delta$, the net influence may either be strong or weak. We conclude that the net influence of $A$ on $C$ is ambiguously positive. So, $++^1 \oplus -^2 = +_?$. Similar observations apply to various other trade-offs.

### 4.3 THE PROPERTY OF SYMMETRY

The sign-propagation algorithm for inference with a regular qualitative network explicitly builds on the properties of symmetry, transitivity, and parallel composition of influences. We have so far addressed the $\otimes$- and $\oplus$-operators and have thereby guaranteed the transitivity and parallel-composition properties of influences. We now focus on the property of symmetry to enable the propagation of qualitative influences over a single arc in the network in both directions.

In a regular qualitative probabilistic network, the property of symmetry guarantees that, if a node $A$ exerts an influence on a node $B$, then node $B$ exerts an influence of the same sign on node $A$. In an enhanced qualitative network, an influence and its reverse also are both positive or both negative. The symmetry property, however, does not hold with regard to the relative magnitudes of an influence and its reverse. The reverse of a strongly positive qualitative influence may be a weakly positive influence, and vice versa. As the relative magnitude of the reverse of a positive influence is unknown, the reverse is taken to be ambiguously positive. A similar observation applies to the reverse of a negative influence.

To conclude, we would like to mention that an alternative way of ensuring that the property of symmetry holds in an enhanced qualitative network is to spec-ify the signs of all reversed influences explicitly; these signs will then have to be elicited from the domain experts involved in the network's construction.

### 4.4 TRADE-OFF RESOLUTION: AN EXAMPLE

In the previous sections, we have argued that the properties of symmetry, transitivity, and parallel composition of influences hold in an enhanced qualitative probabilistic network. The sign-propagation algorithm from Section 2 therefore is generalised straightforwardly to apply to enhanced qualitative networks: instead of the regular $\otimes$- and $\oplus$-operators, it just has to use the enhanced operators for propagating and combining influences. We illustrate the application of the algorithm by means of our running example.

**Example 4.5** We consider once again the qualitative *Antibiotics* network from Figure 3. Suppose that we enter the sign $+$ for node $A$. Node $A$ propagates this sign towards node $T$. Node $T$ thereupon receives the sign $+ \otimes - = -$ and sends it to node $D$. Node $D$ in turn receives the sign $- \otimes + = -$; it does not pass on any sign. Node $A$ also sends its positive sign to node $F$. Node $F$ receives the sign $+ \otimes + = +$ and passes it on to node $D$. Node $D$ then receives the additional sign $+ \otimes + = +$. The two signs that enter node $D$ are combined and result in the ambiguous sign $- \oplus + = ?$.

Now, consider the enhanced *Antibiotics* network from Figure 4. We enter the sign $++^0$ for node $A$; this sign reflects a positive observation for $A$. We once again apply the sign-propagation algorithm, this time using our enhanced operators. Recall that initially all influences' signs have a multiplication-index of 1. Node $A$ propagates its sign towards node $T$. Node $T$ receives the sign $+ +^0 \otimes - -^1 = --^1$ and sends it to node $D$. Node $D$ receives $- -^1 \otimes + +^1 = --^2$. Node $A$ sends its sign $++^0$ also to node $F$. Node $F$ thereupon receives the sign $+ +^0 \otimes+^1 = +^1$ and passes it on to node $D$. Node $D$ receives the additional sign $+^1 \otimes +^1 = +^2$. Combining the two signs that enter node $D$ results in the sign $--^2 \oplus +^2 = -_?$. Note that, while in the regular qualitative network the represented trade-off cannot be resolved and results in an ambiguous influence, the trade-off is resolved in the enhanced qualitative probabilistic network. $\square$

## 5 Conclusions and further research

One of the major drawbacks of qualitative probabilistic networks is their coarse level of detail. Although it may suffice for some problem domains, the coarseness of detail may lead to unresolved trade-offs during inference in other domains. To provide for resolving trade-offs, we have enhanced the formalism of qualitative probabilistic networks by distinguishing between strong and weak influences. We have enhanced the multiplication and addition operators to guarantee the transitivity and parallel-composition properties of influences, thereby generalising the basic sign-propagation algorithm to apply to enhanced qualitative networks. We have shown that our formalism provides for resolving trade-offs in a qualitative, yet efficient way.

Our formalism of enhanced qualitative probabilistic networks does not provide for resolving all possible trade-offs during inference. Since qualitative abstractions do not have the same expressiveness as numerical belief networks, it is hardly likely that any qualitative abstraction will be able to resolve all possible trade-offs. We suspect, however, that in our enhanced signs more information is hidden than we currently exploit upon multiplying and adding influences. In the near future, we will therefore investigate whether still more trade-offs can be resolved within the framework of our enhanced qualitative networks. In addition, we will address the non-associativity of the addition-operator for influences and design heuristics to forestall unnecessary ambiguous results. To conclude, we will extend our formalism to incorporate non-binary variables.